\documentclass[conference]{IEEEtran}
\IEEEoverridecommandlockouts

\usepackage{cite}
\usepackage{amsmath,amssymb,amsfonts}
\usepackage{algorithmic}
\usepackage{graphicx}
\usepackage{epstopdf}
\usepackage{adjustbox}
\usepackage{textcomp}
\usepackage{xcolor}
\usepackage{tabularx}
\def\BibTeX{{\rm B\kern-.05em{\sc i\kern-.025em b}\kern-.08em
    T\kern-.1667em\lower.7ex\hbox{E}\kern-.125emX}}
\begin{document}

\title{TelePlanNet: An AI-Driven Framework for Efficient Telecom Network Planning}

\author{
\IEEEauthorblockN{
Zongyuan Deng\IEEEauthorrefmark{1}\IEEEauthorrefmark{2},
Yujie Cai\IEEEauthorrefmark{2},
Qing Liu\IEEEauthorrefmark{2},
Shiyao Mu\IEEEauthorrefmark{1}\IEEEauthorrefmark{2},
Bin Lyu\IEEEauthorrefmark{1}, and
Zhen Yang\IEEEauthorrefmark{1}
}
\IEEEauthorblockA{
\IEEEauthorrefmark{1}Nanjing University of Posts and Telecommunications, Nanjing 210003, China
}
\IEEEauthorblockA{
\IEEEauthorrefmark{2}China Telecom Corporation Limited Jiangsu Branch, Nanjing 210017, China
}
}

\maketitle

\begin{abstract}
The selection of base station sites is a critical challenge in 5G network planning, which requires efficient optimization of coverage, cost, user satisfaction, and practical constraints. Traditional manual methods, reliant on human expertise, suffer from inefficiencies and are limited to an unsatisfied  planning-construction consistency. Existing AI tools, despite improving efficiency in certain aspects, still struggle to meet the dynamic network conditions and multi-objective needs of telecom operators' networks. To address these challenges, we propose TelePlanNet, an AI-driven framework tailored for the selection of base station sites, integrating a three-layer architecture for efficient planning and large-scale automation. By leveraging large language models (LLMs) for real-time user input processing and intent alignment with base station planning, combined with training the planning model using the improved group relative policy optimization (GRPO) reinforcement learning, the proposed TelePlanNet can effectively address multi-objective optimization, evaluates candidate sites, and delivers practical solutions. Experiments results show that the proposed TelePlanNet can improve the consistency to 78\%, which is superior to the  manual methods, providing telecom operators with an efficient and scalable tool that significantly advances cellular network planning.

\end{abstract}

\section{Introduction}
 The cellular network traffic has maintained a high annual increase over the past three years, escalating the demand for enhanced network capacity. Higher capacity relies on three key factors, i.e., expanded spectrum, higher spectral efficiency, and cell-splitting gains via network densification through base station placement~\cite{Andrews_2024}. For telecom operators, the third factor stands out as a critical topic in network planning, directly addressing escalating traffic demands and coverage requirements in complex, real-world scenarios. Specifically, base stations should be continuously adjusted to accommodate changes in channel transmission conditions driven by urban development. Conventional methods, reliant on human expertise, are slow and with low planning-construction consistency, struggling with poor data processing and limited adaptability to real-world variables.

 To address this above issue, AI tools have been applied to enhance the performance of wireless network planning. For example, methods such as K-means clustering, simulated annealing\cite{zhang2023research} and deep reinforcement learning (RL) \cite{al2024multi} have been explored to reduce deployment costs and improve coverage. However, these tools often oversimplify practical constraints like site availability and user growth trends, requiring extensive manual tuning, which limits their scalability and adaptability in large-scale networks, leading to prolonged planning and the need for adjustments. Large language models (LLMs) \cite{zhao2023survey,xiao2024efficient, wei2022chain} are sophisticated AI systems characterized by their strong semantic understanding and reasoning capabilities. Due to  the powerful semantic understanding and reasoning capabilities, LLMs have also shown broad application prospects in wireless networks, such as empowering wireless intelligence for network planning optimization through fine-tuning and prompt engineering \cite{shao2024wirelessllm}, enabling automated decision-making and user intent understanding in the networking domain \cite{huang2024large},  and managing base stations~\cite{wang2024large}. In \cite{wang2024large}, the authors leveraged LLMs to interpret siting requirements expressed in natural language, automatically construct mathematical optimization models incorporating cost and coverage constraints, and generate executable solver code to determine optimal base station placements. However,  both the dataset and modeling approach in  \cite{wang2024large} struggle to reflect complex real-world telecom scenarios, such as dynamic traffic patterns, terrain constraints, and diverse user demands, thus limiting their adaptability in actual network deployments. Furthermore, LLMs face significant challenges in understanding wireless network data, as they are primarily designed for natural language processing rather than handling numerical and dynamic data in wireless communications. To deal with  this, the work in~\cite{sheng2025beam} adopted extensive data transformation and prompt design, using cross-variable attention mechanisms and prompt-as-prefix techniques. 
 
 Despite these advancements, related work in practical base station siting scenarios still face persistent challenges that hinder effective deployment. First, telecom data is often scattered across systems and contains specific domain characteristics, which limits the training quality for telecom tasks. Second, LLMs process discrete tokens, whereas wireless network performance data is analog, making it challenging to handle such data effectively. Third, LLMs still face difficulties in achieving multi-objective optimization and adapting to dynamic network needs.  These challenges highlight the need for an advanced framework to address the dynamic and multi-objective nature of the selection of base station sites, which motivates the study of this paper. 
 
 In this paper, we propose TelePlanNet, an AI-driven framework to enhance site selection, integrating a three-layer architecture including the data aggregation layer, the intelligent tools layer, and planning execution layer. The data aggregation layer is  responsible for integrating diverse data critical for base station site selection from internal and external sources within telecom operators. The intelligent tools layer serves as the AI-driven core to process  multimodal data and coordinating tasks for base station site selection. While, the planning execution layer translates intelligent analysis
provided by the intelligent tool layer into practical outcomes for planners. In particular, an improved relative policy optimization (GRPO) based RL algorithm with LLMs is desgined for the intelligent tools layer  to reduce the overall training workload and design the reward function. For the design of the reward function,  a staged approach is adopted to evaluate numerically computable features, and LLMs are employed to assess non-quantifiable features.  The proposed TelePlanNet can  bridge planners and network systems, offering a practical 5G site selection solution. Experiment results demonstrate that the proposed TelePlanNet can improve planning-construction consistency to 78\% compared to the historical baseline, which is limited to 70\%.

\section{TelePlanNet Framework Design}
Building upon advancements in AI for base station planning,  we design the TelePlanNet framework to tackle the complex, multi-objective challenges encountered by telecom operators in practical site selection. The framework first employs an LLM to extract semantic information from planning commands, feeding the interpreted planning intent into a planning model designed with the improved GRPO-based RL to generate site selection targets. Through this two-stage decoupling approach, the proposed TelePlanNet effectively addresses the multi-objective optimization challenges inherent in the base station site selection. Additionally, during the training of the GRPO-based RL model, the LLM assists in designing partial reward functions, thereby enhancing the objectivity of the scoring process. This design specifically mitigates the limitation of traditional RL, which typically relies on numerical features in designing reward functions, especially given that base station site selection evaluation involves substantial natural language data.

This section presents the detailed design of the proposed TelePlanNet. As illustrated in Fig. 1, it employs a three-layer architecture consisting of the data convergence layer, intelligent tool layer, and planning execution layer, which integrates an LLM, smart agents, and multi-source data to enhance planning efficiency and resource optimization. In the following, we detail the technical roles and collaborative interactions of each layer.

\begin{figure}[htbp]
    \centering
    \includegraphics[width=1.2\columnwidth, trim=80 80 40 80, clip]{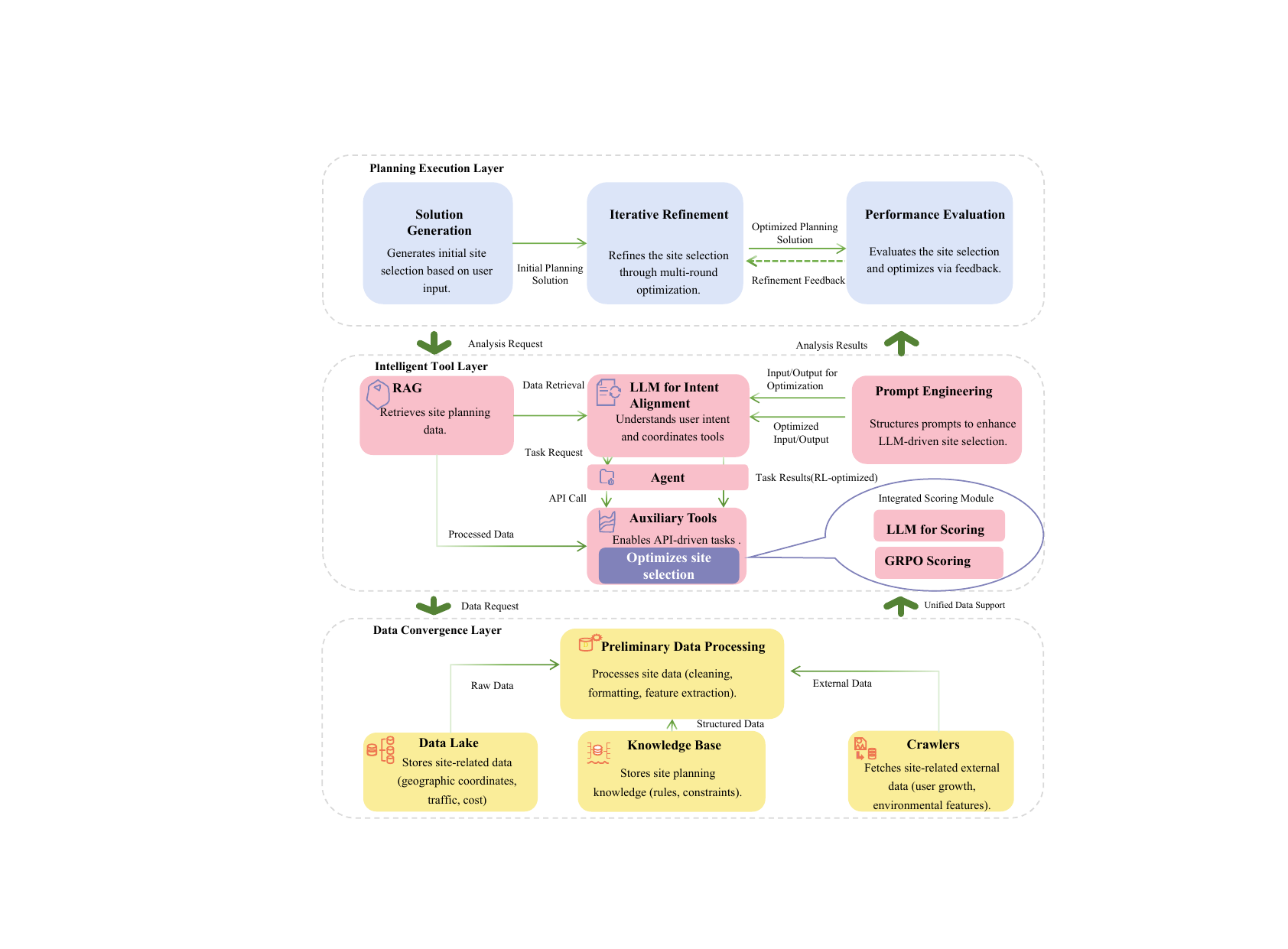}
    \caption{Three-layer architecture of the proposed TelePlanNet.}
\end{figure}
\subsection{Data Aggregation Layer}
The data aggregation layer forms the foundation of the proposed TelePlanNet, tasked with integrating diverse data critical for base station site selection from internal and external sources within telecom operators. This layer integrates network performance, site coordinates, construction costs, and other site-related internal data through a unified data lake. A knowledge is adopted to store site selection rules and constraints. Real-time crawlers are employed to fetch site-related external data, such as user growth trends, new commercial districts, or metro developments. By integrating internal and external data, this layer addresses data fragmentation and delays in data acquisition, ensuring comprehensive and timely site selection decisions. In addition, preprocessing steps including cleaning, formatting, and feature extraction are used to maintain data quality, providing high-quality inputs for the intelligent tool layer.
\subsection{Intelligent Tool Layer}
The intelligent tool layer serves as the AI-driven core of the proposed TelePlanNet, processing multimodal data and coordinating tasks for base station site selection. It encompasses two key technique categories, i.e., intent alignment and tool enhancement.

\textbf{Intent Alignment Techniques}: These leverage the fine-tuned Qwen2.5-32B-Instruct LLM to interpret planners’ intents (e.g., ``improve coverage in a region'') and coordinate associated tools. 
Specifically, fine-tuning  \cite{dodge2020fine} improves understanding of telecom terminology using datasets like TeleQnA~\cite{maatouk2023teleqna} and historical planning data, Zero-Shot enables interpreting of new planning commands, Few-Shot supports sparse-data scenarios, Chain-of-Thought reasoning facilitates multi-objective intent interpreting, and retrieval-augmented generation (RAG) \cite{lewis2020retrieval} enhances planning with site-specific knowledge such as historical site operations, network topology, and policy documents. Additionally, the LLM supports the  GRPO-based RL  model by designing partial reward functions that score non-quantifiable factors, addressing the limitations of numerical feature dependency.

\textbf{Tool Enhancement Techniques}: These enhance the capabilities of the intelligent tool layer in the proposed TelePlanNet by integrating external tools. Specifically, the GRPO-based RL trains the model to enhance site selection decisions under multiple constraints, leveraging quantitative metrics (e.g., coverage and throughput) as rewards, the details of which will be described in Section IV. The LLM provides auxiliary scoring for non-quantifiable factors, with their combined scores guiding optimization decisions. Furthermore, other tools such as K-Means clustering identifies high-density areas for site placement, while ECharts visualizes coverage maps. Smart agents automate workflows via natural language commands, improving site selection efficiency.

\subsection{Planning Execution Layer}
The planning execution layer translates intelligent analysis provided by the intelligent tool l\textbf{}ayer into practical outcomes through an interactive interface for planners. It supports the following functions:
\begin{itemize}
    \item \textbf{Solution generation}: Users input requirements in natural language (e.g., ``improve coverage in an area''), and the system generates initial site selection plans using the intelligent tool layer’s outputs.
    \item \textbf{Iterative refinement}: Users can adjust plans (e.g., ``add base stations within budget while prioritizing complaint data''), and the system updates site selections in real time to meet specific needs, such as increasing the weight of complaint data for targeted improvements.
    \item \textbf{Performance evaluation}: The system compares site selection results with actual construction outcomes and reports, analyzing reasons for discrepancies and generating a semantic report via LLM to support further optimization.
\end{itemize}
Note that the planning execution layer enables fast, intelligent, and practical site selection plans.

\section{Improved GRPO-based RL FOR THE INTELLIGENT TOOL LAYER}

In the process of planning the base station site, operators typically review a wide range of policy documents, industry insight reports, user research reports, and network status analysis. In addition, they  also need to consider whether the chosen base station sites are associated with user complaints. To achieve the base station site planning with the existence of  these challenges, an improved GRPO-based RL algorithm with LLMs is proposed for the intelligent tool layer. In particular, compared to the PPO algorithm, the GRPO algorithm can eliminate the need to train an additional value model and simplifies the calculation of the advantage value, significantly reducing the overall training workload. Moreover, to avoid the limitation that RL computations are typically restricted to numerical features, LLMs are used to extract semantic information from the text and assign the corresponding scores when designing the reward function.

Using the improved GRPO algorithm, our goal is to select the optimal sites from the candidate base station sites to achieve the best network performance at minimal cost. In particular,  a policy \(\pi_\theta\) that selects actions (base station sites) given a state (current set of selected sites and candidate attributes) is optimized to maximize a cumulative reward over a decision trajectory. In the following, we present the details of  the improved GRPO-based RL algorithm for the selection of the base station site, the structure of which is shown in Fig. 2. 

\begin{figure}[htbp]
    \centering
    \includegraphics[width=\columnwidth, trim=50 50 70 30, clip]{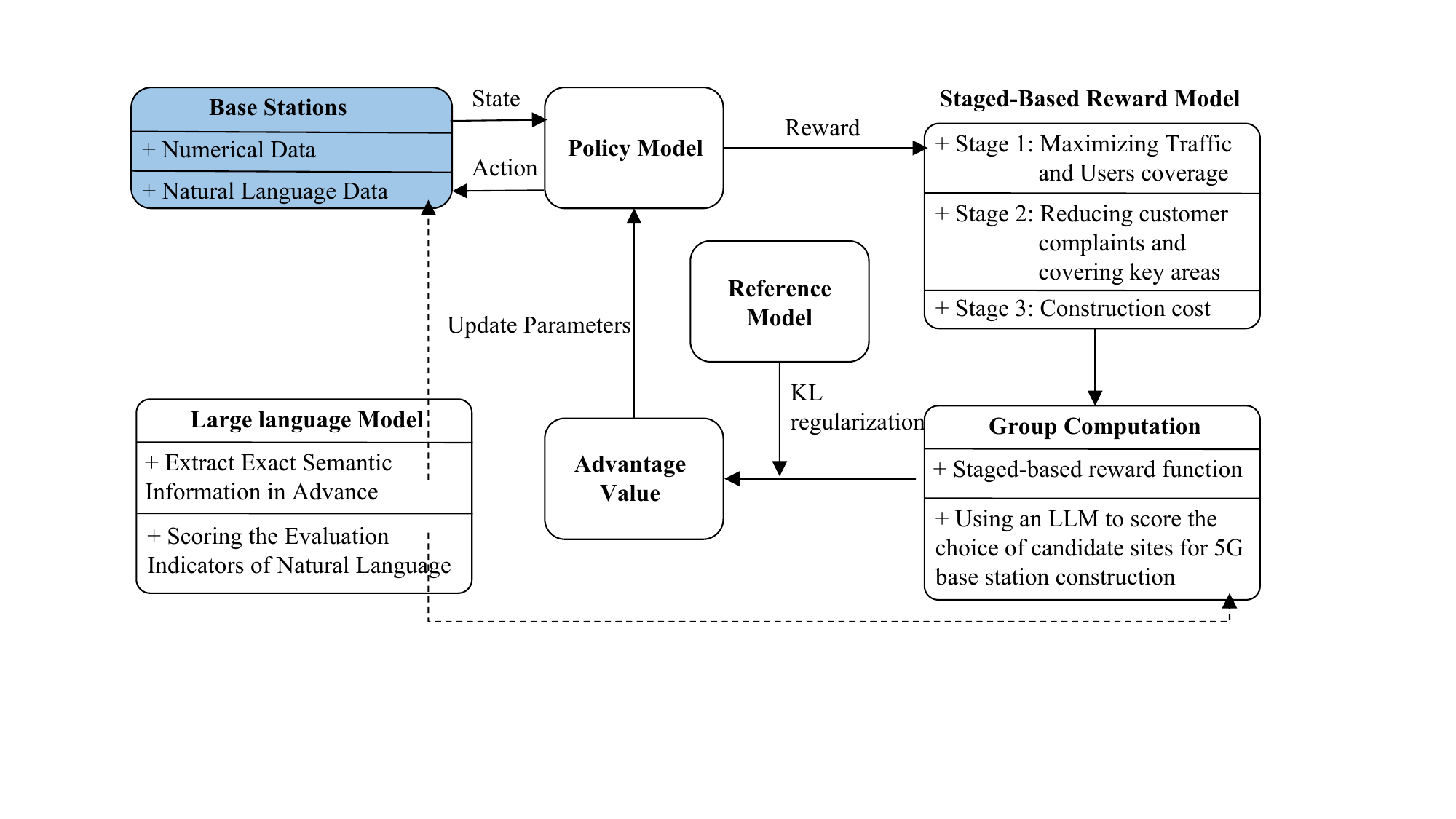}
    \caption{Structure of the improved GRPO-based RL for base station site planning.}
\end{figure}

\subsubsection{Problem Setup}
The base station site planning can be formulated as a Markov Decision Process, which is characterized by the following components:
\begin{itemize}
    \item \textbf{State Space}: The state is composed of the set of base stations already selected along with the attributes of the remaining candidate sites.

    \item \textbf{Action Space}: At each step, a candidate base station is selected from the remaining pool and added to the current one.

    \item \textbf{Transition}: After each action, the state is updated based on the policy model to reflect the new set of selected sites and the revised information of the remaining candidates.

    \item \textbf{Reward}: The reward quantifies the immediate performance impact of selecting a specific base station site, serving as the primary signal to guide the learning process.
\end{itemize}

\subsubsection{GRPO Training Process}

We first establish a Multilayer Perceptron (MLP) as the policy model \(\pi_{\theta}\). This MLP begins with an input layer followed by four fully connected hidden layers, each with 128 hidden units.  Each of these hidden layers employs the Rectified Linear Unit (ReLU) activation function, which helps to introduce non-linearity and enhances the model's ability to learn complex patterns from the input data. Finally, the MLP concludes with an output layer where the softmax activation function is applied, converting the raw output into a probability distribution over the possible actions.

Let \(G\) denote the number of samples that are generated. For each set of candidate base station sets \(c\), the GRPO samples \(G\) different sets of selected base station locations \(\{s_1,s_2,...,s_G\}\) from the old policy \(\pi_{\theta_{\mathrm{old}}}\) and then optimizes the policy model by maximizing the objective \(J{(\theta)}\). \(J{(\theta)}\) is defined as the expected improvement in the performance of the policy relative to its previous behavior, calculated based on relative rewards in groups and regularized by KL divergence constraints to ensure stable updates. Specifically, it can be expressed as 

\begin{equation}
    \begin{aligned}
J{(\theta)} =\;& \mathbb{E}\frac{1}{G}\sum_{i=1}^G \frac{1}{|s_i|} \sum_{t=1}^{|s_i|} \Bigg\{\min\Biggl[
\frac{\pi_\theta(s_{i,t}\mid c,s_{i,<t})}{\pi_{\theta_{\mathrm{old}}}(s_{i,t}\mid c,s_{i,<t})}A_{i,t}, \\
& \text{clip}\Bigl(\frac{\pi_\theta(s_{i,t}\mid c,s_{i,<t})}{\pi_{\theta_{\mathrm{old}}}(s_{i,t}\mid c,s_{i,<t})},1-\epsilon,1+\epsilon\Bigr)A_{i,t}
\Biggr]\\
& -\beta\,\mathbb{D}_{KL}\Bigl[
\pi_{\theta}||\pi_{ref}
\Bigr]\Bigg\},
    \end{aligned}
    \tag{1}
\end{equation}
where \(\epsilon\) and \(\beta\) are hyper-parameters, and \(A_{i,t}\) is the advantage calculated based on relative rewards of the sets of selected base station sites. \(\pi_\theta(s|c)\) is the parameterized policy (e.g., a neural network with parameter \(\theta\)) that outputs the probability of selecting the set \(s\) in the set \(c\). The clip function restricts the probability ratio between the new and old policies to a specific range, i.e., \([1 - \epsilon, 1 + \epsilon]\), thus preventing the new policy from deviating too much from the old policy during an update.  Given a ratio \(r = \frac{\pi_\theta(o_{i,t} \mid q, o_{i,<t})}{\pi_{\text{old}}(o_{i,t} \mid q, o_{i,<t})},\) the clip operator is defined as
\begin{equation}
\label{Clip}
\text{clip}(r, 1-\epsilon, 1+\epsilon) = \max\!\left(\min\!\left(r,\, 1+\epsilon\right),\, 1-\epsilon\right).
    \tag{2}
\end{equation}
\eqref{Clip} indicates that if \(r\) is within the interval \([1-\epsilon,\, 1+\epsilon]\), the value remains unchanged; otherwise, it is set to the nearest bound (either \(1-\epsilon\) or \(1+\epsilon\)). This clipping mechanism is applied in the objective function to restrict the magnitude of policy updates, preventing overly aggressive changes that can destabilize the learning process. \(\mathbb{D}_{KL}\) is a penalty term based on the kullback-leibler (KL) divergence that ensures the current policy \(\pi_{\theta}\) is consistently guided by the reference model \(\pi_{ref}\) during the optimization process.

\subsubsection{Group Computation and Advantage Value}
For each set of base station selection lists \(\{s_1,s_2,...,s_G\}\) generated from the old policy model, we use a reward model to compute the reward values \(R = \{R_1,R_2,...,R_G\}\). For \(i=1,\ldots,G\), \(R_i\) combines two evaluation methods, i.e., stage-based reward function \(r_i\) and an LLM-based score \(\text{LLM\_Score}(s_i)\), which is formulated as \(R_i = w_1 \cdot r_i + w_2 \cdot \text{LLM\_Score}(s_i)\), where \(w_1\) and \(w_2\) are weights. For features that can be computed numerically, we use \(r_i\) defined in \eqref{ri}  to assign rewards; for features that are not convenient to compute numerically, we use an LLM to evaluate and score the planning as \(\text{LLM\_Score}(s_i)\). After normalizing \(R_i\), the advantage value is given by

\begin{equation}
\hat{A}_{i,t} = \tilde{R}_i = \frac{R_i - \operatorname{mean}(\mathbf{R})}{\operatorname{std}(\mathbf{R})}, i =1, \ldots, G,
    \tag{3}
\end{equation}
where  \(\text{mean}(.)\) and \(\operatorname{std}(.) \) denote the operations of calculating the mean and variance, respectively. The advantage value \(\hat{A}_{i,t}\) quantifies how much each selection’s return exceeds the group mean, reducing variance in the policy gradient estimates and guiding updates toward higher‑performing actions.

Considering that there are many factors that influence the selection of base station sites and the importance of these factors varies, the reward function \(r_i\) is divided into three stages to accelerate the convergence of the RL training process. In the first stage, \(r_i\) only considers the most critical factors, i.e.,  throughput and the number of covered users. For example, base station candidates that can support higher traffic
throughput and cover more users receive higher scores. During the second stage, the reward function is enhanced with penalty terms that reflect user complaints metrics and cost factors, such as tower rental fees. In the third stage, recognizing the project objective of contiguous base station deployment, we introduce a geographical clustering score as an incentive term. This biases the model toward selecting multiple adjacent sites within a limited number of regions, thereby promoting concentrated regional coverage. \(r_i\) is formulated as  
\begin{equation}
\label{ri}
r_i = 
\begin{cases} 
w_t t_i + w_u u_i, & \text{Stage 1}, \\
\begin{aligned}
&w_s r_{i_{\text{Stage1}}} - w_m m_i - w_{e} e_i,
\end{aligned} & \text{Stage 2}, \\
w_s r_{i_{\text{Stage2}}} + w_k  k_i, & \text{Stage 3},
\end{cases}
    \tag{4}
\end{equation}
where \(w_t, w_u,w_s, w_m, w_e, w_k\) are weights, \(t_i\) is traffic throughput, \(u_i\) is the number of covered users, \(m_i\) are site complaint issues, \(e_i\) is rental expense, and \(k_i\) is a cluster reward term that measures how tightly the selected sites cluster together. During the training phase, we introduce these features into the reward function in three stages by adding them in batches. 

Note that site complaint issues are typically expressed in natural language, often containing varying degrees of criticism about issues such as poor signal quality or excessive radiation. In the second stage, we extract all these cases from the enterprise's big data lake and use an LLM to classify and perform sentiment grading on these cases. This helps determine how much reward or penalty should be assigned to the construction of a base station at a specific site. Typically, we convert  \(m_i\) into numerical values ranging from \(-10\) to \(10\).
Earlier stage terms are retained at lower weights \(r_{i_{\text{Stage1}}}\) and \(r_{i_{\text{Stage2}}}\) in later stages. After the reward function converges to a stable range at each stage and the model has effectively learned the corresponding objectives, we move on to the next stage.
This staged scheme to reward shaping shown in \eqref{ri} accelerates training convergence while aligning with real-world multi-objective planning priorities. 

\begin{table*}[t]
    \centering
    \renewcommand{\arraystretch}{1.3}
    \caption{Prompt for LLM Evaluation of Base Station Site Selection Score.}
    \begin{tabularx}{\textwidth}{|X|}
        \hline
        \textbf{Instruction:}\\
        Based on the list of base stations marked as ``selected'' in the attachment, identify relevant complaint records, marketing personnel's development demands, and regional development reports for these base station sites. Use this information to evaluate the given 5G base station site selection plan, focusing on the following key aspects:  
        a) Assess the reasonableness of the geographic distribution (e.g., balanced coverage across residential areas, universities, hospitals, etc.).  
        b) Evaluate the satisfaction of policy (e.g., key area coverage requirements).  
        c) Estimate overall user satisfaction (e.g., considering complaint records and user growth forecasts).  
        d) Evaluate whether the plan meets the development demands of marketing personnel.  
        Provide a score between 0 and 10, where 10 indicates the best possible plan and 0 indicates the worst. The GRPO reward function is  evaluated coverage, throughput, complaint rate, and key area coverage, thus focusing on complementary semantic and contextual factors. Provide a brief reasoning for your score. \\
        \textbf{Output Format:}\\
        Score: Score between 0 and 10.  \\
        Reasoning: Brief explanation of the score based on geographic distribution, constraint satisfaction, and user satisfaction.  \\
        \hline
    \end{tabularx}
\end{table*}

However, when evaluating the merits of a base station site selection plan, only computing numerical rewards is insufficient. The key factors, including the development status of the area, the demands of the marketing team, and the coverage of key locations within the area, should be considered. Since these influencing factors are not easily quantifiable, we incorporate a scoring mechanism based on an LLM into the reward function to obtain  \(\text{LLM\_Score}\). The prompt, as shown in Table I, instructs the LLM to retrieve relevant documents about the base station and its surrounding area from the knowledge base and perform an evaluation and scoring. Note that the LLM used currently is the fine-tuned Qwen2.5-32B-Instruct.

\subsubsection{Reference Model}
During the training phase, to accelerate convergence in the improved GRPO-based RL algorithm, we perform supervised fine-tuning (SFT) using recent network construction data based on the policy model \(\pi_\theta\). This pre-trained model also serves as the reference model \(\pi_{ref}\) in the GRPO-based RL algorithm to prevent the policy model from deviating. We calculate the KL divergence between the \(\pi_\theta\) and \(\pi_{ref}\) to ensure that the model's parameter updates do not deviate too much.

\section{EXPERIMENTS AND RESULTS ANALYSIS}
In this section, we  present experiment results to demonstrate the performance of the proposed TelePlanNet for the base station site planning. Without loss of generality, we model the problem as selecting 300 sites from 1,000 candidate base station locations for the deployment of the 5G base station.

\subsection{Dataset Organization and Preprocessing}
Our study uses a data set of 100,000 4G base station sites from telecom operators, one of which has already been upgraded to 5G. For each new 5G deployment phase, a subset of these sites is selected for construction or enhancement. We define the pre-project network status as the initial state and the finalized site list as the optimal solution, incorporating iterative adjustments.

In the  data aggregation layer, amounts of raw data are ingested into a unified data lake and then preprocessed through a centralized data platform. This process includes data cleansing, normalization, and feature extraction to ensure that the dataset remains consistent and is ready for subsequent analysis.

The dataset encompasses 10 phases of 5G projects spanning three years, yielding 500 evaluation records where each record represents a selection of 300 sites chosen from 1,000 candidate locations through contiguous regional planning criteria.
Model performance is evaluated by comparing the generated site selections with actual deployment outcomes, utilizing 400 records allocated for training and 100 reserved for evaluation purposes.
Each site is characterized by multiple features, including basic Information,  user characteristics, base station rental fees, environment features, complaint records, marketer needs, and regional characteristics.

Our dataset is publicly available on Hugging Face at: https://huggingface.co/datasets/CaiYujie/TelePlanNet.

\subsection{Parameter Settings}

For (2),  we set \(\epsilon\) at 0.2 and \(\beta\) at 0.04 to ensure the stability of the update process.
For (4), \(w_s\) is set at 0.2 to ensure that the stage-based reward function does not overlook the rewards from the previous stage. \(w_t\) and \(w_u\) are assigned values of 10 and 12, respectively, to encourage the model to produce solutions that meet the requirements of maximum capacity. \(w_m\), \(w_e\), and \(w_k\) are set to 5, 4, and 8, respectively.

\subsection{Experimental Results}
\begin{figure}[htbp]
    \centering
    \includegraphics[width=\columnwidth]{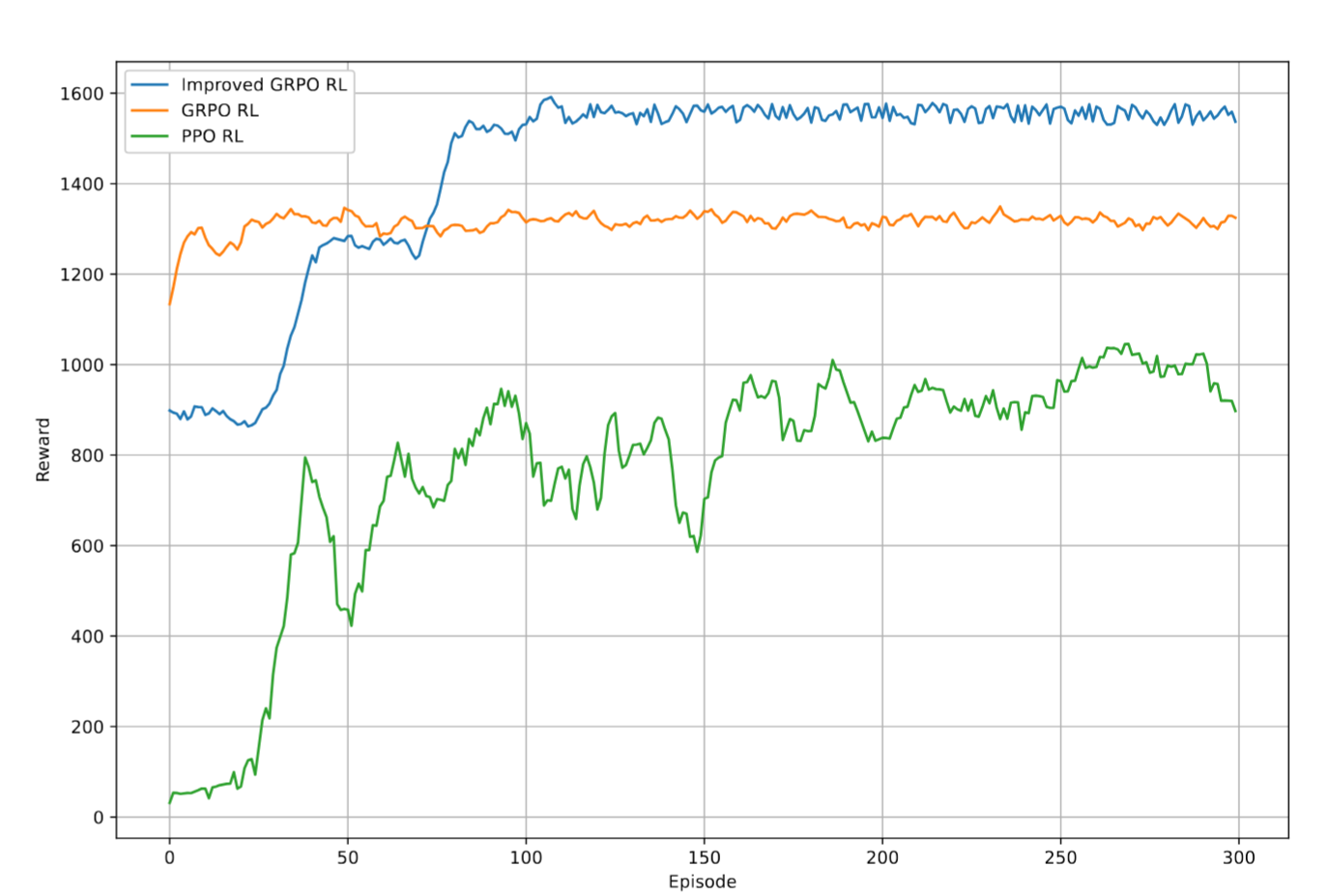}
    \caption{Comparison of reward values among PPO, GRPO, and improved GRPO Algorithms.}
    \label{fig:reward_curve_among_ppo_grpo_and_improved_grpo}
\end{figure}

As shown in Fig. 3, we evaluate the performance of the improved GRPO algorithm with the PPO algorithm and the standard GRPO algorithm. It is observed that the improved GRPO algorithm achieve the largest reward value since this algorithm can gradually introduce sub‑objectives by utilizing  a three‑stage reward scheme described in \eqref{ri}. In particular, this staged scheme helps the agent internalize simple objectives before tackling more complex ones, leading to faster convergence and overall superior network planning strategies.

\begin{figure}[htbp]
    \centering
    \includegraphics[width=\columnwidth]{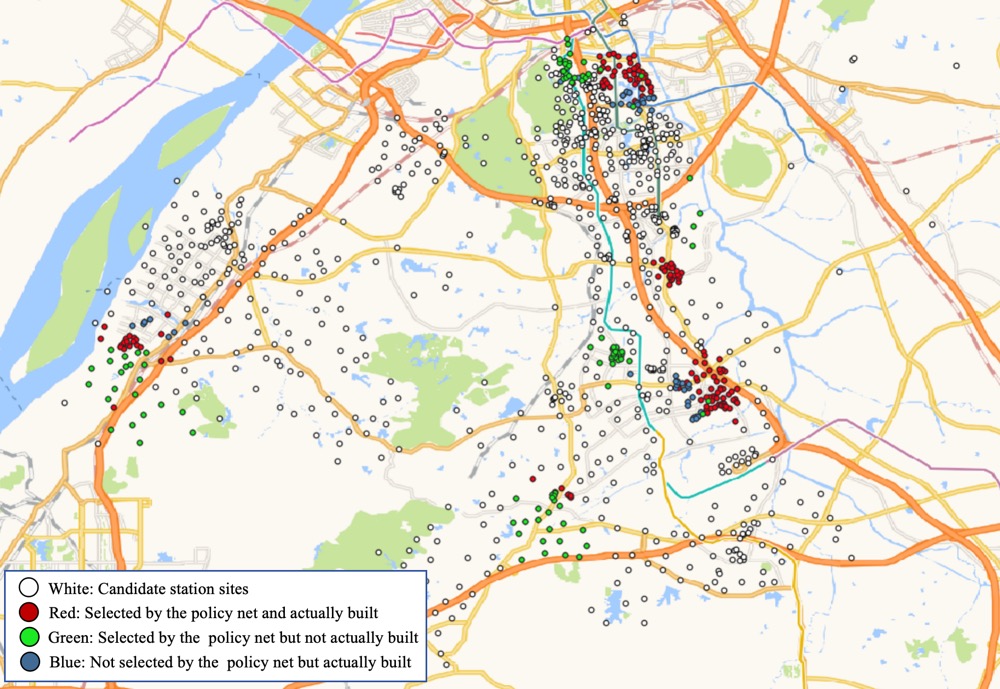}
    \caption{Comparison of base station sites generated using the proposed TelePlanNet and actual constructed sites.}
    \label{fig:site_map}
\end{figure}

In the construction of the 5G base stations, different phases of the project have distinct priorities. The early phases focus on selecting contiguous sites to simplify optimization, while the later phases prioritize coverage completion, reducing the need for contiguous sites. This variation is modeled by adjusting the feature weights in the reward function. In one specific scenario, a high weight on the cluster reward term \(k_i\) emphasizes contiguousness, resulting in site selections concentrated mainly around high-user areas in Jiangning District, Nanjing. Fig. 4 illustrates this result, showing  the sites slected by the proposed TelePlanNet, marked in red and green, with some adjustments that address user complaints about poor signal quality. During construction, initial plans often require modifications, which cause delays. The effectiveness of planning is evaluated by measuring the overlap between the planned and actual built sites, which usually reaches around \(70\%\). The TelePlanNet model achieves an overlap of \(78\%\) with the actual constructions, aligning with the performance of human planners based on historical data.

In order to further evaluate the quality of the  sites selected by the proposed TelePlanNet, we import the list into the  Uetraym, a widely-used simulation software, to analyze the signal strength. We select a contiguous coverage area for signal simulation testing. To simplify the evaluation process, the single channel transmit power is set at 50 W, the antenna downtilt angle is fixed at 10° and the azimuth angle of each base station is configured at 0°, 120° and 240°, respectively. The Uetray evaluates the strength of wireless signals by computing the received power of the reference signal (RSRP). As illustrated in Fig. 5, every grid within the simulation area exhibits an RSRP value above –80 dBm, and over 60 percent of the grids exceed –60 dBm. Consequently, only those lakeside sectors with comparatively weaker coverage require targeted blind-spot compensation. This shows that the  base station sites selected by the TelePlanNet achieves contiguous coverage while delivering enhanced signal strength.

\begin{figure}[htbp]
    \centering
    \includegraphics[width=\columnwidth]{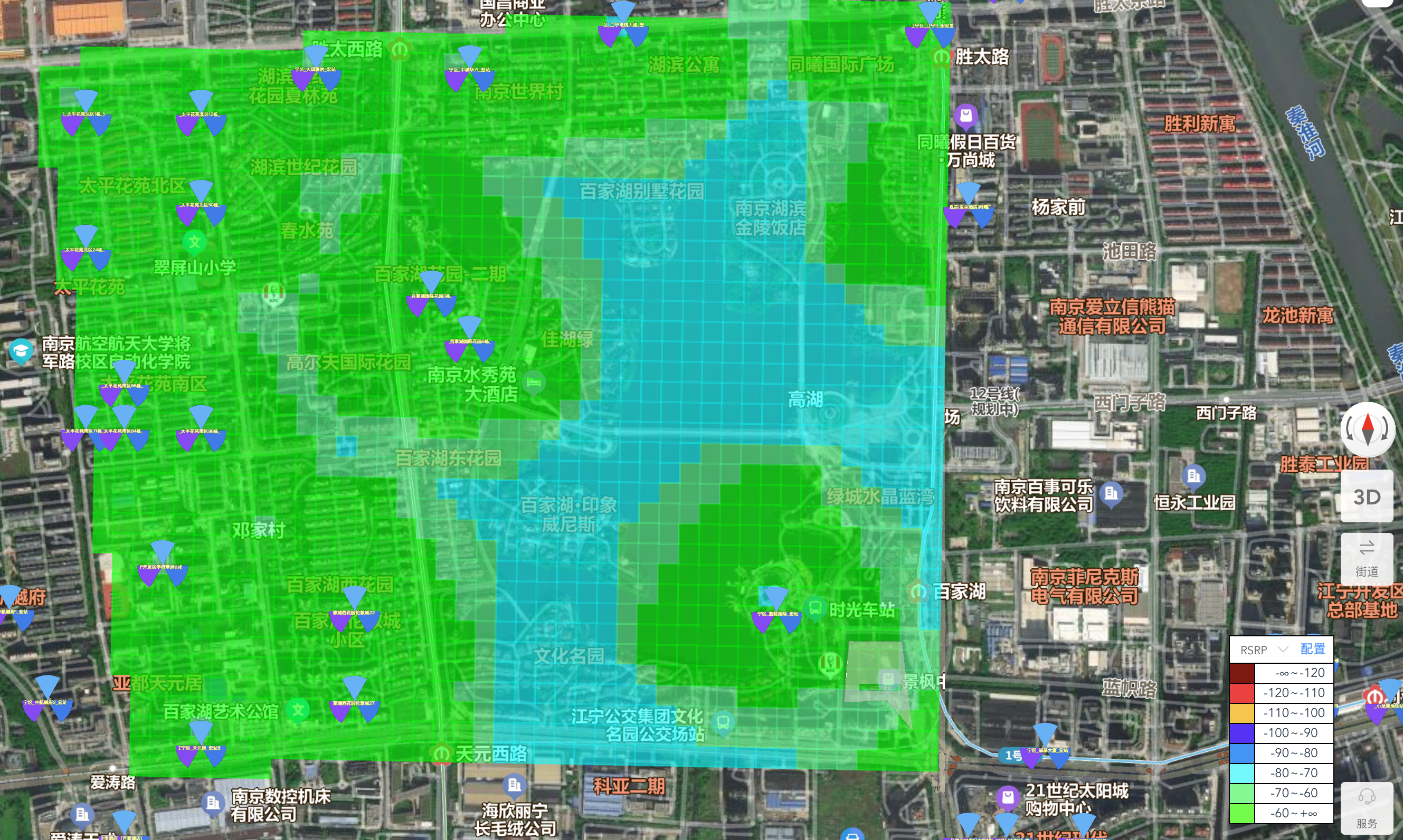}
    \caption{Signal intensity analysis of the sites selected by the TelePlanNet.}
    \label{fig:simulation_map}
\end{figure}

In terms of improving the efficiency of wireless base station site planning, we define the planning time as the duration from user input to the final scheme. Historical manual planning takes an average of 95 hours per 1,000 sites, while TelePlanNet reduces it to 62 hours, achieving a 35\% reduction in planning time.

\section{CONCLUSION}
This paper has proposed TelePlanNet, an operator-centric framework tailored for real-world telecommunication networks to support human-AI collaboration and allow quantitative evaluation of planning solutions, aiming to  improve the efficiency and adaptability of network planning. To generate optimal site selection strategies, the improved GRPO-based RL with LLMs has been designed.  The experiment results demonstrate that the proposed TelePlanNet can improve the consistency of planning and construction to 78\% compared to the historical baseline, which is restricted to the consistency of 70\%. Moreover, the proposed TelePlanNet  can achieve a 35\% reduction in planning
time compared to the manual planning method.


\vspace{12pt}

\end{document}